\documentclass{article} 
\usepackage[T1]{fontenc}
\usepackage[preprint]{colm2026_conference}

\usepackage{microtype}
\usepackage{hyperref}
\usepackage{url}
\usepackage{booktabs}
\usepackage{amsmath}
\usepackage{amssymb}
\usepackage{algorithm}
\usepackage{algorithmic}
\usepackage{graphicx}
\usepackage{multirow}
\usepackage{wrapfig}
\usepackage{subcaption}
\usepackage{pgfplots}
\pgfplotsset{compat=1.18}
\definecolor{pal1}{HTML}{B2B0F1} 
\definecolor{pal2}{HTML}{9B99EA}
\definecolor{pal3}{HTML}{827FE4}
\definecolor{pal4}{HTML}{6863D6}
\definecolor{pal5}{HTML}{4E46C4} 
\definecolor{voteBase}{HTML}{9B99EA} 
\definecolor{voteHigh}{HTML}{4E46C4} 

\definecolor{ablBase}{HTML}{90CAF9}  
\definecolor{ablHigh}{HTML}{1565C0}  

\usepackage{lineno}

\definecolor{darkblue}{rgb}{0, 0, 0.5}
\hypersetup{colorlinks=true, citecolor=darkblue, linkcolor=darkblue, urlcolor=darkblue}

\title{Beyond Symbolic Solving: Multi Chain-of-Thought Voting for Geometric Reasoning in Large Language Models\vspace{-2mm}}

\author{
  Md. Abu Bakor Siddique$^\dagger$\hfill Shahrin Hossain$^\dagger$\hfill  
  Sadman Ahmed Siam$^\dagger$\\
  \quad\quad\hspace{-1pt}\textbf{Syed Rifat Raiyan}$^\ddagger$\quad\hfill  
  \hspace{-7pt}\textbf{Hasan Mahmud}\hfill  
  \textbf{Md Kamrul Hasan}\quad\hfill \\
  \hfil Systems and Software Lab (SSL), Department of Computer Science and Engineering\hfil \\
  \hfil Islamic University of Technology, Dhaka, Bangladesh\hfil \\
  \vspace{0.5mm} 
  \fontsize{9pt}{10pt}\selectfont 
  \hfil $^\dagger$Equal contribution \quad $^\ddagger$Corresponding author: \texttt{rifatraiyan@iut-dhaka.edu}\hfil 
}

\begin{document}

\ifcolmsubmission
\linenumbers
\fi

{\centering
\maketitle}

\vspace{-6mm}
\begin{abstract}
\vspace{-2mm}
%
%
%
%
Geometric Problem Solving (GPS) remains at the heart of enhancing mathematical reasoning in large language models because it requires the combination of diagrammatic understanding, symbolic manipulation and logical inference. In existing literature, researchers have chiefly focused on synchronising the diagram descriptions with text literals and solving the problem. In this vein, they have either taken a neural, symbolic or neuro-symbolic approach. But this solves only the first two of the requirements, namely diagrammatic understanding and symbolic manipulation, while leaving logical inference underdeveloped. The logical inference is often limited to one chain-of-thought (CoT). To address this weakness in hitherto existing models, this paper proposes \textbf{MARS-GPS}, that generates multiple parallel reasoning rollouts augmented with Python code execution for numerical verification, ranks them using token-level entropy as a confidence signal, and aggregates answers through a multi-stage voting and self-verification pipeline. Empirical results show that \textbf{MARS-GPS} with 8 parallel rollouts achieves \textbf{88.8\%} on Geometry3K, a nearly \textbf{+11\%} improvement over the prior state-of-the-art, with accuracy scaling consistently as the number of rollouts increases from 1 to 16 
(+6.0\% on ablation subset). We provide our code and data in an anonymous repository: \url{https://anonymous.4open.science/r/MARS-GPS-DE55}.
\end{abstract}

\vspace{-3mm}
\section{Introduction}
\label{sec:introduction}
\vspace{-1mm}







Geometry Problem Solving is regarded as one of the pinnacles of human reasoning. In short, GPS takes a diagram and a textual description, and attempts to solve a problem. There are essentially two steps in solving the problem. Initial task is to identify the given knowledge base, \textit{i.e.}, analysing the diagram and note down what is given. If the diagram does not contain enough information beyond the basic shapes, it is required that the diagram be annotated with proper information from textual description.

The second task is to utilise the theorems to derive further conclusions.  The prime difficulty in this case arises in identifying relevant theorems. For instance, Pythagoras' theorem will not be applicable to circle-related problems. Some problems do not might not have an exclusive solution path, making matters complicated.

Models such as Pi-GPS \citep{Zhao_2025_ICCV}, MINT-CoT \citep{chen2025mintcotenablinginterleavedvisual}, PGPSNet-v2 \citep{zhang2024fusereasonverifygeometry}, G-LLaVA \citep{gao2025gllavasolvinggeometricproblem}, etc., have attempted to perfect diagrammatic understanding, that is, the former part of the process. But, as we shall show, concentrating on logical inference through multiple Chain-of-Thoughts(CoTs) can significantly improve performance in GPS. Our contributions can be summarised as follows: 

\begin{itemize}
   \item Showing that parallel rollout sampling outperforms symbolic solvers
   \item Introducing a training-free confidence signal derived from per-token log probabilities at zero additional cost
   \item An aggregation algorithm combining majority voting, entropy ranking, and Large Language Models self-verification 
   \item State-of-the-art results on Geometry3K and PGPS9K
\end{itemize}
\section{Related Work}
\label{sec:related_work}

Researchers in GPS have focused on a few specific approaches. Symbolic solvers such as InterGPS \citep{lu2021intergpsinterpretablegeometryproblem} attempt to solve the problem through logical manipulation. These systems are have limited scalability, motivating the development of neuro-symbolic solvers. Neuro-symbolic solvers such as PGPSNet \citep{ijcai2023p376}, PGPSNet-v2 \citep{zhang2024fusereasonverifygeometry}, DualGeoSolver \citep{xiao2024learningsolvegeometryproblems}, FormalGeo \citep{zhang2024formalgeoextensibleformalizedframework} mix both neural networks and symbolic solvers to solve the geometry problems. This approach is more scalable than purely symbolic systems., but due to difficulty handling complex reasoning chains and reliance on theorem sets, these models ultimately fail to get better at reasoning. Finally, Multimodal Large Language Models-based approaches such as G-LLaVA \citep{gao2025gllavasolvinggeometricproblem}, GeoUni \citep{cheng2025geouniunifiedmodelgenerating} treat GPS as a multimodal reasoning task, ultimately suffering from the similar kind of unreliable reasoning. Most recent works include the likes of \citet{Zhao_2025_ICCV}, who propose Pi-GPS that uses diagrams to disambiguate textual formal language via a rectifier-verifier micro module. This approach highlights the importance of diagram information in GPS, but relies on theorem sets provided. Similarly, \citet{chen2025mintcotenablinginterleavedvisual} propose MINT-CoT which interleaves fine-grained visual tokens into chain-of-thought reasoning steps via an Interleave Token mechanism. This approach highlights the benefits of visual grounding during reasoning, but the reasoning is still unreliable.

Inference-time scaling is performance enhancement during inference. \citet{balachandran2025inference} show that inference-time scaling can improve mathematical reasoning, but has less success with geometry as GPS requires more multimodal reasoning, and inference-time scaling remains more applicable in text-heavy setting. As this paper will show, their insight is also relevant for GPS.


Benchmarks such as Geometry3k, MathVista, MathVerse, GeoEval are used to evaluate performance of the GPS systems. The problem sets need to be carefully selected so that both diagram and the text have equal significance, so that it can be checked how much they are being accurately parsed by the system and solved. 
\section{Method}
\label{sec:method}

\subsection{Problem Formulation}
\label{sec:problem_formulation}

We consider the standard geometry problem solving (GPS) setting: given a
natural-language problem description $T$ and an accompanying diagram image
$I$, the goal is to generate the correct answer $a \in \mathcal{A}$, where
$\mathcal{A} = \{A, B, C, D\}$ represents the set of multiple-choice candidates.
 
We use the structured representation from Pi-GPS~\citep{Zhao_2025_ICCV}, where
$(T, I)$ is parsed into a set of first-order geometric predicates $\mathcal{F}$
(\textit{e.g.}, $\texttt{Perpendicular}(\text{Line}(B,D), \text{Line}(D,C))$). We
refer to the tuple $(\mathcal{F}, \mathcal{A})$ as the problem instance's
\textit{structured context}. 
 
Our approach operates entirely at inference time which means no model weights
are adjusted or fine-tuned. We use two axes to scale test-time computation. First,
a frozen large language model $f_\theta$ is given access to a \textit{code
execution sandbox} $\mathcal{E}$ which is a live Python kernel that runs code
created by $f_\theta$ in the middle of reasoning and injects the actual output back into its
context. Second, we generate $k$ independent solution attempts in parallel,
each producing a candidate answer $\hat{a}_m \in \mathcal{A} \cup \{\emptyset\}$.
The final answer $\hat{a}$ is selected from $\{\hat{a}_1, \ldots, \hat{a}_k\}$
via a \textit{parallel voting} strategy described in
Section~\ref{sec:verification}.

\subsection{Diagram Understanding and Parsing}
\label{sec:diagram_understanding}
 
Since frontier MLLMs have difficulty extracting precise logical relationships
directly from geometry diagrams, we
translate both modalities through a two-stage parsing pipeline into a
unified formal representation $\mathcal{F}^*$, which is used as input to our
inference-time reasoning strategy (see Section~\ref{sec:reasoning_strategy}).
 
\paragraph{Text Parser.}
The text parser applies a rule-based regular-expression pipeline to $T$,
producing formal literals $\mathcal{F}_T$ such as
$\texttt{Find}(\texttt{AreaOf}(\texttt{Triangle}(A,B,C)))$ or
$\texttt{Equals}(\texttt{LengthOf}(\texttt{Line}(A,B)), 13)$.
We prefer this design over neural alternatives because geometry datasets
are relatively small, and the downstream reasoning stage is highly
sensitive to malformed formalizations. Empirically, the rule-based parser
achieves 97\% accuracy, compared with 67\% for a BART-based baseline
\citep{lu2021intergpsinterpretablegeometryproblem}.

\paragraph{Diagram Parser.}
The diagram $I$ is processed with PGDPNet ~\citep{Zhang2022}, which
extracts geometric primitives and their relations as formal literals
$\mathcal{F}_D$ (\textit{e.g.}, $\texttt{PointLiesOnLine}(D, \texttt{Line}(B,C))$,
$\texttt{Equals}(\texttt{LengthOf}(\texttt{Line}(A,B)), 5)$).

The final representation $\mathcal{F}^*$ merges text literals
$\mathcal{F}_T$ and diagram literals $\mathcal{F}_D$ into a unified structured
description of each problem and serves as the \textit{only} input passed to
$f_\theta$ at inference time.

\begin{figure*}[t]
    \centering
    \includegraphics[width=\textwidth]{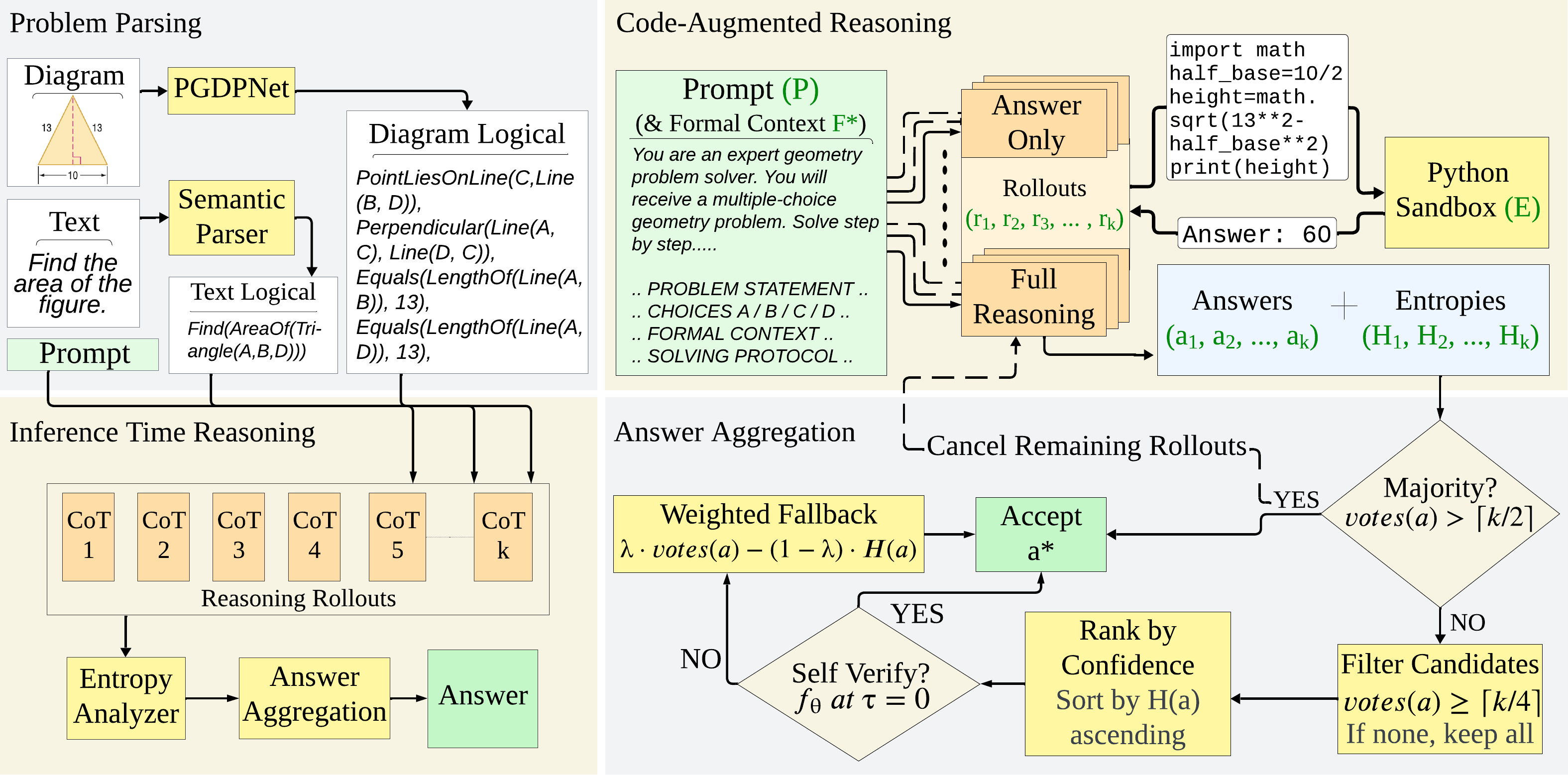}
    \caption{Overview of the \textbf{\underline{M}}ulti-path \textbf{\underline{A}}ggregated \textbf{\underline{R}}easoning \textbf{\underline{S}}ystem for \textbf{\underline{G}}eometry \textbf{\underline{P}}roblem \textbf{\underline{S}}olving (\textbf{MARS-GPS}) pipeline. \textit{Left:} the problem 
parsing stage takes the diagram and problem text as input and produces a unified 
formal context $\mathcal{F}^*$ via PGDPNet and a rule-based semantic parser. 
\textit{Right:} the inference-time ensemble reasoning stage samples $k$ parallel 
rollouts from $f_\theta$, each augmented with a Python sandbox $\mathcal{E}$ for 
numerical computation. The rollout outputs feed into the answer aggregation 
pipeline, which applies majority voting, entropy-ranked self-verification, and 
a weighted fallback to produce the final answer $a^*$.}
    \label{fig:pipeline}
\end{figure*}
 
\subsection{Inference-Time Reasoning Strategy}
\label{sec:reasoning_strategy}
 
Prior neural-symbolic approaches to GPS invest their computation budget at
training time, learning theorem predictors \citep{lu2021intergpsinterpretablegeometryproblem},
reinforcement-trained search policies \citep{peng2023geodrl}, or fine-tuned
multimodal encoders \citep{ijcai2023p376}. At inference time, these methods
commit to a single reasoning path: one symbolic execution, one answer. This
single-path paradigm is a wrong step which propagates irrecoverably to
an incorrect answer.
 
We take a different approach. Rather than relying on a deterministic symbolic
solver, we use $f_\theta$ (GPT-OSS 120B, served via
VLMs \citep{10.1145/3600006.3613165}) to reason directly over $\mathcal{F}^*$, and
instead of committing to a single trace, we sample $k$ independent reasoning
rollouts in parallel and aggregate their outputs using a confidence-aware
selection strategy. The three components of this approach are described below:
parallel rollout sampling (Section~\ref{sec:rollouts}), confidence estimation
via token entropy (Section~\ref{sec:entropy}), and code-augmented reasoning
(Section~\ref{sec:code}).
 
\subsubsection{Parallel Rollout Sampling}
\label{sec:rollouts}
 
Given $\mathcal{F}^*$, we construct a structured prompt $\mathcal{P}$ with a
\textit{system prompt} instructing $f_\theta$ to output $\backslash\texttt{boxed\{N\}}$
where $N \in \{1,2,3,4\}$, and a \textit{user prompt} containing $T$,
$\mathcal{A}$, and $\mathcal{F}^*$. The raw image $I$ is never passed to
$f_\theta$ as all visual information is already encoded in $\mathcal{F}^*$.
 
We sample $k$ independent rollouts in parallel:
\begin{equation}
    \{r_1, r_2, \ldots, r_k\} \sim f_\theta(\mathcal{P} \mid \mathcal{F}^*)
\end{equation}
Each rollout $r_i$ is a complete chain-of-thought trace terminating with a
boxed answer $a_i$, extracted via pattern matching on $\backslash\texttt{boxed\{N\}}$.
Rollouts with no extractable answer are excluded from aggregation. The $k$
rollouts run concurrently via a thread pool of 16 workers, with VLMs'
PagedAttention batching all requests in a single forward pass making
wall-clock time for $k=8$ only marginally greater than $k=1$. Each problem
has a maximum budget of 900 seconds \citep{wang2023selfconsistency}.
 
\subsubsection{Confidence Estimation via Token Entropy}
\label{sec:entropy}
 
VLMs return per-token log probabilities $\ell_{t,j} = \log p_\theta(w_j \mid w_{<t})$
at no additional cost. At each token position $t$ in rollout $r_i$, we compute
Shannon entropy over the top-$v$ vocabulary entries:
\begin{equation}
    H_t = -\sum_{j} e^{\ell_{t,j}} \cdot \log_2\!\left(e^{\ell_{t,j}}\right)
\end{equation}
and aggregate into a per-rollout mean entropy:
\begin{equation}
    \bar{H}_i = \frac{1}{T_i} \sum_{t=1}^{T_i} H_t
    \label{eq:mean_entropy}
\end{equation}
$\bar{H}_i$ serves as an inverse confidence score which means lower is more confident.
It is used to break vote ties and to rank candidates for self-verification,
ensuring the most confident answer is verified first
(Section~\ref{sec:verification}).
 
\subsubsection{Code-Augmented Reasoning}
\label{sec:code}
 
Geometry problems often require precise numerical computation that Large Language Models handle
unreliably through pure token generation~\citep{wang2024mathcoder}. Each
rollout $r_i$ is therefore paired with a sandbox instance $\mathcal{E}$:
\begin{equation}
    r_i = f_\theta(\mathcal{P},\ \mathcal{E}), \quad
    \mathcal{E}: \text{code} \mapsto \text{output}
\end{equation}
When $f_\theta$ writes a Python code block mid-reasoning, it is executed in
$\mathcal{E}$ and the output is injected back into context. We maintain a pool
of 16 persistent $\mathcal{E}$ instances to avoid kernel startup overhead while
keeping rollouts isolated. In practice, near 40\% of rollouts invoke $\mathcal{E}$ at least once,
with usage concentrated on computationally intensive problems
requiring precise numerical calculation. The full procedure is detailed in
Algorithm~\ref{alg:reasoning} (see Appendix~\ref{sec:appendix_algorithms}).
 
\subsection{Verification and Self-Consistency}
\label{sec:verification}
 
Having generated $k$ rollouts with answers $\{a_i\}_{i=1}^k$ and confidence
scores $\{\bar{H}_i\}_{i=1}^k$, the remaining challenge is to reliably pick
the correct answer from this pool. Na\"ive majority voting is a natural
baseline, but it treats all rollouts as equally trustworthy. We instead use a
six-step procedure that progressively filters candidates using vote counts,
entropy scores, and self-verification, falling back to weighted scoring only
when stronger signals are unavailable.
 
\paragraph{Step 1: Early Consensus.}
We first check whether any answer appears in $k/2 + 1$ or more of the $k$ rollouts.
If so, we accept it immediately:
\begin{equation}
    \text{if } \sum_{i=1}^{k} \mathbf{1}[a_i = a] \geq {k/2+1}
    \quad \Rightarrow \quad a^* = a
\end{equation}
Near-unanimous agreement across independent rollouts is a strong correctness
signal, and this early exit avoids spending verification budget on easy cases.
 
\paragraph{Step 2: Hard Accept.}
If no answer reaches $\lceil k/2 \rceil + 1$ votes, we check for a weaker consensus of $\lceil k/2 \rceil$ or
more. For example: an answer supported by four of eight rollouts constitutes an absolute
majority and is accepted directly. In case of a tie, we move on to step 3.
 
\paragraph{Step 3: Candidate Selection.}
If neither condition is met, we collect the \textit{candidate answers} or
those appearing in $k/4$ or more rollouts:
\begin{equation}
    \mathcal{A}_{\text{cand}} = \left\{ a \;\middle|\;
    \sum_{i=1}^{k} \mathbf{1}[a_i = a] \geq \lceil k/4 \rceil \right\}
\end{equation}
Single-rollout answers are treated as outliers and discarded. If
$\mathcal{A}_{\text{cand}}$ is empty (all answers are singletons), all four
choices are retained.
 
\paragraph{Step 4: Entropy-Ranked Verification.}
For each candidate $a \in \mathcal{A}_{\text{cand}}$, we compute its mean
support entropy across the rollouts that produced it:
\begin{equation}
    \bar{H}(a) = \frac{1}{|\mathcal{I}_a|}
    \sum_{i \in \mathcal{I}_a} \bar{H}_i,
    \quad \mathcal{I}_a = \{i : a_i = a\}
\end{equation}
Candidates are sorted in ascending order of $\bar{H}(a)$, most confident
first and submitted to self-verification in this order. Verifying the most
confident candidate first is efficient: if it passes, we stop without querying
the model for remaining candidates.
 
\paragraph{Step 5: Large Language Models Self-Verification.}
For each candidate $a$ (in entropy-ranked order), we query $f_\theta$ at
temperature $\tau = 0$ with a structured verification prompt asking whether
the proposed answer is \texttt{CORRECT} or \texttt{WRONG}. The deterministic
setting ensures stable, reproducible decisions. If $f_\theta$ responds
\texttt{CORRECT}, we accept $a$ as $a^*$ and terminate; if \texttt{WRONG}, we
move to the next candidate. This step exploits the model's own reasoning to
cross-check candidates against $\mathcal{F}^*$, without requiring a separately
trained verifier.
 
\paragraph{Step 6: Weighted Fallback.}
If all candidates are rejected, we fall back to a scoring function combining
vote count and confidence:
\begin{equation}
    a^* = \arg\max_{a \in \mathcal{A}_{\text{cand}}} \;
    \lambda \cdot \text{votes}(a) \;-\;
    (1-\lambda) \cdot \bar{H}(a)
    \label{eq:fallback}
\end{equation}
where $\bar{H}(a)$ is subtracted
because lower entropy should be rewarded. This fallback is triggered in fewer
than 8\% of problems, indicating that earlier steps resolve the vast majority
of cases. The full aggregation procedure is detailed in 
Algorithm~\ref{alg:aggregation} (see Appendix~\ref{sec:appendix_algorithms}).
 
\subsection{Full Pipeline Summary}
\label{sec:pipeline}
 
Figure~\ref{fig:pipeline} illustrates the complete system. Our approach
decomposes GPS into two sequential stages with a clean interface between them.
 
\textbf{Stage 1 --- Problem Parsing}
(Section~\ref{sec:diagram_understanding}) takes the raw problem $(T, I)$ as
input and produces the formal representation $\mathcal{F}^*$. This stage
involves two components: a rule-based text parser producing $\mathcal{F}_T$,
and the PGDPNet diagram parser producing $\mathcal{F}_D$. All visual
understanding happens here, the raw image $I$ is never forwarded to the
reasoning model.
 
\textbf{Stage 2 --- Inference-Time Ensemble Reasoning}
(Sections~\ref{sec:reasoning_strategy}--\ref{sec:verification}) takes
$\mathcal{F}^*$ as input and produces the final answer $a^*$. This is our
primary contribution, consisting of three components: parallel rollout
sampling from the frozen $f_\theta$, per-rollout confidence estimation via
token entropy, and confidence-aware aggregation with Large Language Models self-verification.
 
We refer to our full system as \textbf{MARS-GPS}. Table~\ref{tab:main}
reports results against all prior neural-symbolic baselines on Geometry3K
and PGPS9K.

\section{Experimental Setup}
\label{sec:experimental_setup}

\subsection{Datasets}
We have run the experiments on Geomety3k and PGPS9K datasets. Geometry3K is made of 3002 geometric problems. It has 3 splits. There are 2101 problems to train, 300 problems is used for validation and finally 601 datapoints for testing purpose. Each of the data points have a problem statement, geometric diagram, and formal language parsing annotaions. 

PGPS9K is an expanded version of Geometry3K that contains 9022 datapoints. It has 4000 unique diagrams. Of these, 2891 problems are taken directly from the Geometry3K dataset and the rest are from high school text books. 

Collectively, they cover almost all types of plane geometry problems that can be found in high school textbooks.


\subsection{Baselines}
We are using GPT-OSS (120B)~\citep{openai2025gptoss120bgptoss20bmodel} at the heart of our pipeline. We also used state of the art AI models and analyze their performance in geometry problem solving as detailed below.

\textbf{Neural Methods.}
We use NGS ~\citep{chen2021geoqa}, which encodes diagrams with ResNet-101, Geoformer ~\citep{chen2022unigeo}, which uses VL-T5 for diagram encoding, SCA-GPS ~\citep{ning2023scagps}, a symbolic-character-aware model, GOLD ~\citep{zhang2024gold}, which converts diagrams to natural language descriptions, PGPSNet-v2-S ~\citep{zhang2024fusereasonverifygeometry}, which combines CNN and GRU encoders with a fuse-reason-verify pipeline, and LANS ~\citep{li2024lans}, a layout-aware neural solver that relies on ground-truth diagram annotations.

\textbf{Neural-Symbolic Methods.}
We compare against Inter-GPS ~\citep{lu2021intergpsinterpretablegeometryproblem}, which parses problems into formal language and applies symbolic theorem search, GeoDRL ~\citep{peng2023geodrl}, which extends Inter-GPS with reinforcement learning for theorem prediction, E-GPS ~\citep{wu2024egps}, which combines top-down solving with bottom-up program generation, and Pi-GPS ~\citep{Zhao_2025_ICCV}, which uses a rectifier-verifier module to disambiguate text using diagram information.

\textbf{Multimodal Large Language Models (MLLMs).}
For MLLMs evaluated as direct multimodal solvers, we include GPT-4o ~\citep{openai2024gpt4ocard}, Gemini~2 ~\citep{gemini2023}, Claude~3.5 Sonnet ~\citep{anthropic2024claude35} and Qwen-VL ~\citep{bai2023qwenvl}, all processing both problem text and diagrams end-to-end.

\textbf{Proprietary Large Language Models.}
We additionally compare against proprietary Large Language Models evaluated on parsed formal representations: GPT-5 ~\citep{openai2025gpt5}, GPT-5.2 ~\citep{zhang2026pritpg} and Claude~4.5 Sonnet ~\citep{anthropic2025claude45}. Results are reported in Table ~\ref{tab:main}.

All results are reported using top-1 accuracy: the percentage of problems for which the system's final answer matches the ground-truth label. Since both benchmarks are multiple-choice, this is equivalent to exact-match accuracy over $\mathcal{A} = \{A, B, C, D\}$.



\section{Results}
\label{sec:results}

\subsection{Main Results}
\label{sec:main_results}
\begin{table}[t]
\begin{center}
\setlength{\tabcolsep}{4pt}
\begin{tabular}{@{}llcc@{}}
\toprule
\textbf{Category} & \textbf{Method} & \textbf{Geometry3K} & \textbf{PGPS9K} \\
\midrule
\multirow{4}{*}{MLLMs}
 & Qwen-VL \citep{bai2023qwenvl} & 26.7 & 23.2 \\
 & GPT-4o \citep{openai2024gpt4ocard} & 58.6 & 51.0 \\
 & Claude 3.5 Sonnet \citep{anthropic2024claude35} & 56.4 & 45.9 \\
 & Gemini 2 \citep{gemini2023} & 60.7 & 56.8 \\
\midrule
\multirow{3}{*}{\shortstack[l]{Proprietary\\LLMs }}
 & GPT-5 \citep{openai2025gpt5} & 61.5 & -- \\
 & GPT-5.2 \citep{zhang2026pritpg} & 73.1 & -- \\
 & Claude 4.5 Sonnet \citep{anthropic2025claude45} & 75.8 & -- \\
\midrule
\multirow{6}{*}{Neural Methods}
 & NGS \citep{chen2021geoqa} & 58.8 & 46.1 \\
 & Geoformer \citep{chen2022unigeo} & 59.3 & 47.3 \\
 & SCA-GPS \citep{ning2023scagps} & 76.7 & -- \\
 & GOLD$^*$ \citep{zhang2024gold} & 62.7 & 60.6 \\
 & PGPSNet-v2-S$^*$ \citep{zhang2024fusereasonverifygeometry} & 76.4 & 69.2 \\
 & LANS (Diagram GT)$^*$ \citep{li2024lans} & 82.3 & 74.0 \\
\midrule
\multirow{4}{*}{\shortstack[l]{Neural-symbolic\\Methods}}
 & Inter-GPS \citep{lu2021intergpsinterpretablegeometryproblem} & 57.5 & -- \\
 & GeoDRL \citep{peng2023geodrl} & 68.4 & 66.7 \\
 & E-GPS \citep{wu2024egps} & 67.9 & -- \\
 & Pi-GPS \citep{Zhao_2025_ICCV} & 77.8 & 69.8 \\
\midrule
 & \textbf{MARS-GPS (ours)} & \textbf{88.8} & \textbf{77.48} \\
\bottomrule
\end{tabular}
\end{center}
\caption{Comparison on Geometry3K and PGPS9K. Best results in \textbf{bold}. $^*$ indicates models trained on the larger PGPS9K dataset.}
\label{tab:main}
\end{table}
We have put our results comparing with other baselines in table~\ref{tab:main}. We can see that MARS-GPS consistently outperforms all other benchlines on Geometry3k dataset by a significant margine. Even in the cases where the Multimodal Large Language Models have apparent better performance, they lack precision with geometric nuances. The table captures this phenomenon. Generalized Large Language Models face difficulty in accurately parsing precise geometric figures and computing values embedded in them accurately, such as the length of a side of a square or value of degree in a triangle. Our method yields higher accuracy by leveraging superior reasoning capabilities of Large Language Models while using a dedicated diagram parser and correction methods. 

Against Neural-symbolic Methods, our model has an notable  \textbf{11\%} improvement over frontier works like PI-GPS and an even more noteworthy \textbf{30\%} over pioneering works like Inter-GPS in this domain. These results effectively solidify our claim that using superior reasoning capabilites of Large Language Models yield better results than using rule based Neuro-symbolic approach.

We present comparisons on PGPS9k dataset to further solidify the effectiveness of our method. MARS-GPS achieves \textbf{77.48\%} accuracy which is nearly \textbf{8\%} higher than PI-GPS and over \textbf{20\%} more than general purpose Multimodal Large Language Models. These results demonstrate the comprehensive nature of our pipeline and robustness against different types of geometric problems.

We have also presented comparisons against top performing Neural models, Notably LANS, which was trained on larger PGPS9K dataset. It achieves \textbf{82\%} accuracy thanks to its superior focus in diagram parsing and ground truth annotations on train data. However, it still gets outperformed by MARS-GPS. 





\subsection{Ablation Studies}
\label{sec:ablation}

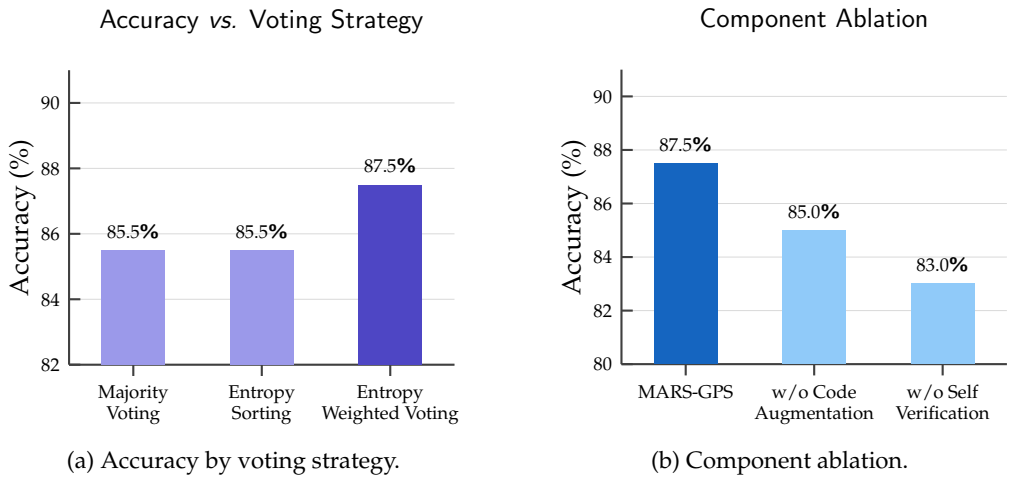
\begin{figure}[t]
    \centering
    \begin{subfigure}[t]{0.48\textwidth}
        \centering
        \begin{tikzpicture}
            \begin{axis}[
                width=\linewidth, 
                height=5.5cm,
                ybar,
                bar width=24pt, 
                xmin=0.5, xmax=3.5,
                ymin=82, ymax=91,
                ytick={82, 84, 86, 88, 90},
                ylabel={Accuracy (\%)},
                title={\textsf{Accuracy \textit{vs.} Voting Strategy}},
                title style={yshift=1ex},
                ylabel style={yshift=-1ex},
                xtick={1, 2, 3},
                xticklabels={Majority\\Voting, Entropy\\Sorting, Entropy\\Weighted Voting},
                xticklabel style={align=center, font=\scriptsize}, 
                yticklabel style={font=\scriptsize},               
                axis lines*=left,
                ymajorgrids=true,
                grid style={line width=.3pt, draw=gray!30},
                axis line style={thick, black!75},
                tick style={thick, black!75},
                nodes near coords={\pgfmathprintnumber[fixed, fixed zerofill, precision=1]{\pgfplotspointmeta}\%},
                nodes near coords style={font=\sffamily\bfseries\scriptsize, text=black, yshift=1pt},
            ]
            \addplot[fill=voteBase, draw=none, bar shift=0pt] coordinates {(1, 85.5)};
            \addplot[fill=voteBase, draw=none, bar shift=0pt] coordinates {(2, 85.5)};
            \addplot[fill=voteHigh, draw=none, bar shift=0pt] coordinates {(3, 87.5)};
            \end{axis}
        \end{tikzpicture}
        \caption{Accuracy by voting strategy.}
        \label{fig:voting_strategies}
    \end{subfigure}
    \hfill
    \begin{subfigure}[t]{0.48\textwidth}
        \centering
        \begin{tikzpicture}
            \begin{axis}[
                width=\linewidth, 
                height=5.5cm,
                ybar,
                bar width=24pt,
                xmin=0.5, xmax=3.5,
                ymin=80, ymax=91, 
                ytick={80, 82, 84, 86, 88, 90},
                ylabel={Accuracy (\%)},
                title={\textsf{Component Ablation}},
                title style={yshift=1ex},
                ylabel style={yshift=-1ex},
                xtick={1, 2, 3},
                xticklabels={MARS-GPS, w/o Code\\Augmentation, w/o Self\\Verification},
                xticklabel style={align=center, font=\scriptsize}, 
                yticklabel style={font=\scriptsize},               
                axis lines*=left,
                ymajorgrids=true,
                grid style={line width=.3pt, draw=gray!30},
                axis line style={thick, black!75},
                tick style={thick, black!75},
                nodes near coords={\pgfmathprintnumber[fixed, fixed zerofill, precision=1]{\pgfplotspointmeta}\%},
                nodes near coords style={font=\sffamily\bfseries\scriptsize, text=black, yshift=1pt},
            ]
            \addplot[fill=ablHigh, draw=none, bar shift=0pt] coordinates {(1, 87.5)};
            \addplot[fill=ablBase, draw=none, bar shift=0pt] coordinates {(2, 85.0)};
            \addplot[fill=ablBase, draw=none, bar shift=0pt] coordinates {(3, 83.0)};
            \end{axis}
        \end{tikzpicture}
        \caption{Component ablation.}
        \label{fig:component_ablation}
    \end{subfigure}
    \caption{Ablation studies on a subset of Geometry3K. (a)~Entropy-weighted voting outperforms majority voting and entropy sorting by 2.0 percentage points. (b)~Removing self-verification causes the largest single-component accuracy drop ($-4.5$pp), followed by code augmentation ($-2.5$pp).}
    \label{fig:ablation}
\end{figure}

\textbf{Voting Strategies} We compare three voting strategies for aggregating predictions across sampled reasoning chains. Majority voting ~\citep{wang2023selfconsistency} selects the most frequently occurring answer and achieves an accuracy of 85.5\% on our evaluation set. Entropy sorting ranks candidate answers by their mean token entropy and selects the lowest-entropy prediction, also yielding 85.5\%. Entropy-weighted voting extends majority voting by weighting each candidate answer by the inverse of its entropy, thereby down-weighting uncertain predictions; this strategy achieves the highest accuracy of 87.5\%, outperforming both baselines by 2 percentage points (see Figure~\ref{fig:voting_strategies}). Based on these results, entropy-weighted voting is used as the default aggregation strategy in MARS-GPS.

\textbf{Code-augmented reasoning.} We augment each reasoning rollout with access to a sandboxed Python executor, allowing the model to offload precise numerical computation and reducing arithmetic hallucinations~\citep{wang2024mathcoder}. Removing code augmentation drops accuracy from 87.5\% to 85.0\% (Figure~\ref{fig:component_ablation}), a 2.5 percentage point decrease. We observe that 42\% of all CoT rollouts invoke code execution at least once; among those rollouts, denying code access causes accuracy to fall to 75\%, indicating that the problems requiring computation are disproportionately harder and benefit most from symbolic verification. Detailed
per-problem breakdown is provided in
Appendix~\ref{sec:appendix_sandbox}. \\

\textbf{Self-verification.} The self-verification stage prompts the model to re-examine its solution for logical consistency before committing to a final answer. As shown in Figure~\ref{fig:component_ablation}, removing self-verification reduces accuracy from 87.5\% to 83.0\%. It is a 4.5 percentage point performance drop, the largest single-component degradation. This confirms that self-verification acts as an effective post-hoc filter, catching theorem misapplications and cascading calculation errors that survive the initial reasoning pass. \\ 

\begin{wrapfigure}{r}{0.40\textwidth}
    \centering
    \vspace{-4.5mm}
    \includegraphics[width=0.38\textwidth]{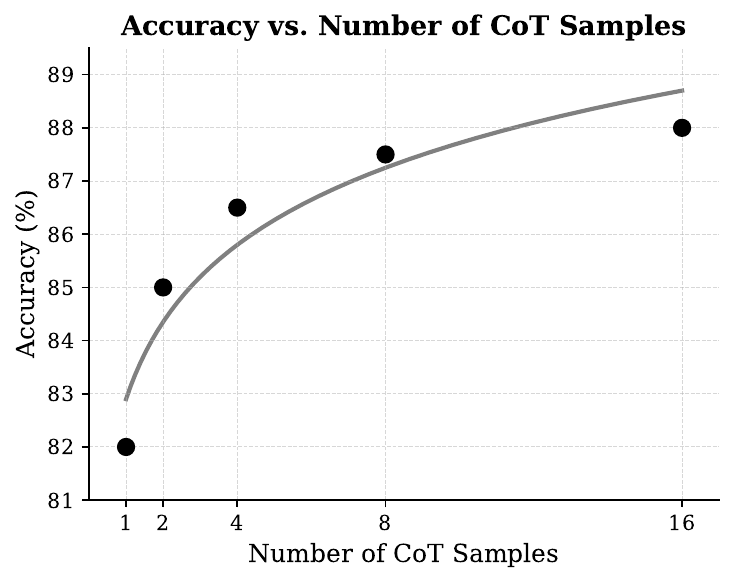}
    \caption{Accuracy \textit{vs.}\ number of CoT samples.}
    \vspace{-5mm}
    \label{fig:cot_scaling}
\end{wrapfigure}

\textbf{Accuracy \textit{vs.}\ number of CoT samples.}
To establish the relationship between accuracy gain and the number of CoTs,
we ran tests for $k \in \{1, 2, 4, 8, 16\}$ CoT samples on Geometry3K.
As shown in Figure~\ref{fig:cot_scaling}, accuracy increases log-linearly with
the number of samples, rising from 82.0\% ($k{=}1$) to
88.0\% ($k{=}16$), consistent with the log-linear scaling behaviour observed
in self-consistency decoding~\citep{wang2023selfconsistency}. We observe
diminishing returns between $k{=}8$ and $k{=}16$, where doubling the
compute budget yields only a 0.5 percentage point improvement. Based on this, the main
experiments use $k{=}8$ to balance accuracy and computational overhead.
Detailed per-$k$ results are provided in
Appendix~\ref{sec:appendix_full_results}.

\smallskip
\noindent{\footnotesize\textit{Note:} We conduct ablation experiments on a randomly sampled subset of Geometry3K to isolate the contribution of each pipeline component. Unless stated otherwise, all ablations use $k{=}8$ rollouts with entropy-weighted voting.}





\section{Analysis and Discussion}
\label{sec:analysis}

\paragraph{Error Analysis.}
We inspected incorrectly answered problems from the Geometry3K test set to identify failure modes. The hardest category turns out to be area-based problems with 77.4\% accuracy. This result is consistent with Pi-GPS findings on area ambiguity. The second hardest category is the determination of length/other parameters with 85.6\%. Other categories such as trigonometry, circle, angle and quadrilateral problems don't pose much difficulty to the system. 

Another point to note is that the wrong predictions take almost 4.2 times more execution time than correct ones, as can be seen from the table~\ref{tab:timing_summary} of Appendix E. The probable reason is because the rollouts disagree, entropy is high, and the system exhausts the verification steps before falling back. Circle and Length/Other problems take the longest time before predicting wrong answers. This means that the system spends its full verification budget on these cases before falling back, as shown in the table~\ref{tab:timing_category} of the Appendix E.

\paragraph{When Does Inference-Time Scaling Help?}

Parallel rollout sampling provides the largest gains on problems of moderate difficulty those requiring 3--5 reasoning steps. On simple problems (1--2 steps), a single rollout already succeeds, so additional samples add little. On the hardest problems (6+ steps, auxiliary constructions), all rollouts tend to fail in similar ways, and voting cannot recover from a systematically wrong approach. This pattern is consistent with observations by \citet{balachandran2025inference} on other mathematical domains, and suggests that combining inference-time scaling with training-time improvements to the reasoning model could push accuracy further on the hardest tier.

\paragraph{Limitations.}

Our approach inherits the limitations of its parsing stage: problems where PGDPNet produces an incomplete or incorrect $\mathcal{F}^*$ cannot be recovered by downstream reasoning, regardless of how many rollouts are sampled. Additionally, the computational cost scales linearly with $k$; while VLMs' batching makes this practically efficient, it still represents a $k$-fold increase in token generation over a single-pass baseline. Finally, MARS-GPS is currently limited to multiple-choice geometry problems and has not been evaluated on open-ended or proof-based tasks. This might be a potential avenue for researchers working in the autoformalisation domain, to combine these strategies in proving theorems using lean or CoT.

\section{Future Work}
\label{sec:future_work}
Looking ahead, there are several promising directions. First, improving the diagram parsing stage, either through better neural parsers or by incorporating Multimodal Large Language Models directly into parsing, could address the largest single source of errors. Second, combining inference-time scaling with training-time improvements, such as fine-tuning $f_\theta$ on geometry-specific data, may yield compounding gains. Third, extending the framework to open-ended geometry problems and formal theorem proving would broaden its applicability. Finally, adaptive rollout budgets or sampling fewer rollouts for easy problems and more for hard ones could reduce computational cost without sacrificing accuracy.

\section{Conclusion}
\label{sec:conclusion}

In this paper, we propose MARS-GPS, an inference-time framework for geometry problem solving. It generates multiple parallel reasoning rollouts, estimates their confidence through token-level entropy, and finally aggregates answers via a multi-stage voting and self-verification pipeline. Without any training or fine-tuning, MARS-GPS achieves \textbf{88.8\%}  on Geometry3K and \textbf{77.5\%} on PGPS9K, outperforming the existing approaches. Our ablation studies confirm that accuracy scales log-linearly with the number of rollouts and that entropy-weighted voting is the most effective aggregation strategy.




\clearpage

\bibliography{colm2026_conference}

@InProceedings{Zhao_2025_ICCV,
    author    = {Zhao, Junbo and Zhang, Ting and Sun, Jiayu and Tian, Mi and Huang, Hua},
    title     = {Pi-GPS: Enhancing Geometry Problem Solving by Unleashing the Power of Diagrammatic Information},
    booktitle = {Proceedings of the IEEE/CVF International Conference on Computer Vision (ICCV)},
    month     = {October},
    year      = {2025},
    pages     = {1526-1536}
}

@article{balachandran2025inference,
  title   = {Inference-Time Scaling for Complex Tasks: Where We Stand and What Lies Ahead},
  author  = {Balachandran, Vidhisha and Chen, Jingya and Chen, Lingjiao and Garg, Shivam and Joshi, Neel and Lara, Yash and Langford, John and Nushi, Besmira and Vineet, Vibhav and Wu, Yue and Yousefi, Safoora},
  journal = {CoRR},
  volume  = {abs/2504.00294},
  year    = {2025},
  doi     = {10.48550/arXiv.2504.00294},
  url     = {https://arxiv.org/abs/2504.00294},
  eprint  = {2504.00294},
  archivePrefix = {arXiv},
  primaryClass  = {cs.CL}
}

@inproceedings{Zhang2022,
  title = {Plane Geometry Diagram Parsing},
  author = {Zhang, Ming-Liang and Yin, Fei and Hao, Yi-Han and Liu, Cheng-Lin},
  booktitle = {Proceedings of the Thirty-First International Joint Conference on Artificial Intelligence, {IJCAI-22}},
  pages = {1636--1643},
  year = {2022},
  month = {7},
  doi = {10.24963/ijcai.2022/228}
}

@inproceedings{ijcai2023p376,
  title     = {A Multi-Modal Neural Geometric Solver with Textual Clauses Parsed from Diagram},
  author    = {Zhang, Ming-Liang and yin, Fei and Liu, Cheng-Lin},
  booktitle = {Proceedings of the Thirty-Second International Joint Conference on
               Artificial Intelligence, {IJCAI-23}},
  publisher = {International Joint Conferences on Artificial Intelligence Organization},
  editor    = {Edith Elkind},
  pages     = {3374--3382},
  year      = {2023},
  month     = {8},
  note      = {Main Track},
  doi       = {10.24963/ijcai.2023/376},
  url       = {https://doi.org/10.24963/ijcai.2023/376},
}

@inproceedings{10.1145/3600006.3613165,
author = {Kwon, Woosuk and Li, Zhuohan and Zhuang, Siyuan and Sheng, Ying and Zheng, Lianmin and Yu, Cody Hao and Gonzalez, Joseph and Zhang, Hao and Stoica, Ion},
title = {Efficient Memory Management for Large Language Model Serving with PagedAttention},
year = {2023},
isbn = {9798400702297},
publisher = {Association for Computing Machinery},
address = {New York, NY, USA},
url = {https://doi.org/10.1145/3600006.3613165},
doi = {10.1145/3600006.3613165},
booktitle = {Proceedings of the 29th Symposium on Operating Systems Principles},
pages = {611–626},
numpages = {16},
location = {Koblenz, Germany},
series = {SOSP '23}
}

@misc{chen2025mintcotenablinginterleavedvisual,
      title={MINT-CoT: Enabling Interleaved Visual Tokens in Mathematical Chain-of-Thought Reasoning}, 
      author={Xinyan Chen and Renrui Zhang and Dongzhi Jiang and Aojun Zhou and Shilin Yan and Weifeng Lin and Hongsheng Li},
      year={2025},
      eprint={2506.05331},
      archivePrefix={arXiv},
      primaryClass={cs.CV},
      url={https://arxiv.org/abs/2506.05331}, 
}

@misc{zhang2024fusereasonverifygeometry,
      title={Fuse, Reason and Verify: Geometry Problem Solving with Parsed Clauses from Diagram}, 
      author={Ming-Liang Zhang and Zhong-Zhi Li and Fei Yin and Liang Lin and Cheng-Lin Liu},
      year={2024},
      eprint={2407.07327},
      archivePrefix={arXiv},
      primaryClass={cs.AI},
      url={https://arxiv.org/abs/2407.07327}, 
}

@misc{gao2025gllavasolvinggeometricproblem,
      title={G-LLaVA: Solving Geometric Problem with Multi-Modal Large Language Model}, 
      author={Jiahui Gao and Renjie Pi and Jipeng Zhang and Jiacheng Ye and Wanjun Zhong and Yufei Wang and Lanqing Hong and Jianhua Han and Hang Xu and Zhenguo Li and Lingpeng Kong},
      year={2025},
      eprint={2312.11370},
      archivePrefix={arXiv},
      primaryClass={cs.CL},
      url={https://arxiv.org/abs/2312.11370}, 
}

@inproceedings{
wang2024mathcoder,
title={MathCoder: Seamless Code Integration in {LLM}s for Enhanced Mathematical Reasoning},
author={Ke Wang and Houxing Ren and Aojun Zhou and Zimu Lu and Sichun Luo and Weikang Shi and Renrui Zhang and Linqi Song and Mingjie Zhan and Hongsheng Li},
booktitle={The Twelfth International Conference on Learning Representations},
year={2024},
url={https://openreview.net/forum?id=z8TW0ttBPp}
}

@misc{lu2021intergpsinterpretablegeometryproblem,
      title={Inter-GPS: Interpretable Geometry Problem Solving with Formal Language and Symbolic Reasoning}, 
      author={Pan Lu and Ran Gong and Shibiao Jiang and Liang Qiu and Siyuan Huang and Xiaodan Liang and Song-Chun Zhu},
      year={2021},
      eprint={2105.04165},
      archivePrefix={arXiv},
      primaryClass={cs.CL},
      url={https://arxiv.org/abs/2105.04165}, 
}

@misc{xiao2024learningsolvegeometryproblems,
      title={Learning to Solve Geometry Problems via Simulating Human Dual-Reasoning Process}, 
      author={Tong Xiao and Jiayu Liu and Zhenya Huang and Jinze Wu and Jing Sha and Shijin Wang and Enhong Chen},
      year={2024},
      eprint={2405.06232},
      archivePrefix={arXiv},
      primaryClass={cs.AI},
      url={https://arxiv.org/abs/2405.06232}, 
}

@misc{zhang2024formalgeoextensibleformalizedframework,
      title={FormalGeo: An Extensible Formalized Framework for Olympiad Geometric Problem Solving}, 
      author={Xiaokai Zhang and Na Zhu and Yiming He and Jia Zou and Qike Huang and Xiaoxiao Jin and Yanjun Guo and Chenyang Mao and Yang Li and Zhe Zhu and Dengfeng Yue and Fangzhen Zhu and Yifan Wang and Yiwen Huang and Runan Wang and Cheng Qin and Zhenbing Zeng and Shaorong Xie and Xiangfeng Luo and Tuo Leng},
      year={2024},
      eprint={2310.18021},
      archivePrefix={arXiv},
      primaryClass={cs.AI},
      url={https://arxiv.org/abs/2310.18021}, 
}

@misc{cheng2025geouniunifiedmodelgenerating,
      title={GeoUni: A Unified Model for Generating Geometry Diagrams, Problems and Problem Solutions}, 
      author={Jo-Ku Cheng and Zeren Zhang and Ran Chen and Jingyang Deng and Ziran Qin and Jinwen Ma},
      year={2025},
      eprint={2504.10146},
      archivePrefix={arXiv},
      primaryClass={cs.LG},
      url={https://arxiv.org/abs/2504.10146}, 
}

@misc{anthropic2024claude35,
  title={Claude 3.5 Sonnet},
  author={{Anthropic}},
  year={2024},
  howpublished={\url{https://www.anthropic.com/news/claude-3-5-sonnet}}
}

@article{gemini2023,
  title={Gemini: A Family of Highly Capable Multimodal Models},
  author={{Google Gemini Team}},
  journal={arXiv preprint arXiv:2312.11805},
  year={2023}
}

@article{bai2023qwenvl,
  title={Qwen-VL: A Versatile Vision-Language Model for Understanding, Localization, Text Reading, and Beyond},
  author={Bai, Jinze and Bai, Shuai and Yang, Shusheng and Wang, Shijie and Tan, Sinan and Wang, Peng and Lin, Junyang and Zhou, Chang and Zhou, Jingren},
  journal={arXiv preprint arXiv:2308.12966},
  year={2023}
}

@inproceedings{chen2021geoqa,
  title={GeoQA: A Geometric Question Answering Benchmark Towards Multimodal Numerical Reasoning},
  author={Chen, Jiaqi and Tang, Jianheng and Qin, Jinghui and Liang, Xiaodan and Liu, Lingbo and Xing, Eric and Lin, Liang},
  booktitle={Findings of the Association for Computational Linguistics: ACL-IJCNLP 2021},
  pages={513--523},
  year={2021}
}

@inproceedings{chen2022unigeo,
  title={UniGeo: Unifying Geometry Logical Reasoning via Reformulating Mathematical Expression},
  author={Chen, Jiaqi and Li, Tong and Qin, Jinghui and Lu, Pan and Lin, Liang and Chen, Chongyu and Liang, Xiaodan},
  booktitle={Proceedings of the 2022 Conference on Empirical Methods in Natural Language Processing},
  pages={3313--3323},
  year={2022}
}

@inproceedings{ning2023scagps,
  title={A Symbolic Characters Aware Model for Solving Geometry Problems},
  author={Ning, Maizhen and Wang, Qiu-Feng and Huang, Kaizhu and Huang, Xiaowei},
  booktitle={Proceedings of the 31st ACM International Conference on Multimedia},
  pages={7767--7775},
  year={2023}
}

@inproceedings{zhang2024gold,
  title={GOLD: Geometry Problem Solver with Natural Language Description},
  author={Zhang, Jiaxin and Moshfeghi, Yashar},
  booktitle={Findings of the Association for Computational Linguistics: NAACL 2024},
  pages={263--278},
  year={2024}
}

@inproceedings{li2024lans,
  title={LANS: A Layout-Aware Neural Solver for Plane Geometry Problem},
  author={Li, Zhong-Zhi and Zhang, Ming-Liang and Yin, Fei and Liu, Cheng-Lin},
  booktitle={Findings of the Association for Computational Linguistics: ACL 2024},
  pages={2596--2608},
  year={2024}
}

@inproceedings{peng2023geodrl,
  title={GeoDRL: A Self-Learning Framework for Geometry Problem Solving Using Reinforcement Learning in Deductive Reasoning},
  author={Peng, Shuai and Fu, Di and Liang, Yijun and Gao, Liangcai and Tang, Zhi},
  booktitle={Findings of the Association for Computational Linguistics: ACL 2023},
  pages={13468--13480},
  year={2023}
}

@inproceedings{wu2024egps,
  title={E-GPS: Explainable Geometry Problem Solving via Top-Down Solver and Bottom-Up Generator},
  author={Wu, Wenjun and Zhang, Lingling and Liu, Jun and Tang, Xi and Wang, Yaxian and Wang, Shaowei and Wang, Qianying},
  booktitle={Proceedings of the IEEE/CVF Conference on Computer Vision and Pattern Recognition (CVPR)},
  pages={13828--13837},
  year={2024}
}

@article{openai2025gpt5,
  title={GPT-5 Technical Report},
  author={{OpenAI Team}},
  journal={arXiv preprint},
  year={2025}
}

@article{zhang2026pritpg,
  title={Prior-Guided Multi-Step Theorem Prediction via Theorem Precedence Graphs},
  author={Zhang, S. and others},
  journal={arXiv preprint arXiv:2603.04852},
  year={2026}
}

@article{anthropic2025claude45,
  title={Claude 4.5 Model Card},
  author={{Anthropic Team}},
  journal={arXiv preprint arXiv:2511.19773},
  year={2025}
}

@article{wang2023selfconsistency,
  title     = {Self-Consistency Improves Chain of Thought Reasoning in Language Models},
  author    = {Wang, Xuezhi and Wei, Jason and Schuurmans, Dale and Le, Quoc and Chi, Ed and Narang, Sharan and Chowdhery, Aakanksha and Zhou, Denny},
  journal   = {International Conference on Learning Representations (ICLR)},
  year      = {2023}
}

@misc{openai2025gptoss120bgptoss20bmodel,
      title={gpt-oss-120b \& gpt-oss-20b Model Card}, 
      author={OpenAI and : and Sandhini Agarwal and Lama Ahmad and Jason Ai and Sam Altman and Andy Applebaum and Edwin Arbus and Rahul K. Arora and Yu Bai and Bowen Baker and Haiming Bao and Boaz Barak and Ally Bennett and Tyler Bertao and Nivedita Brett and Eugene Brevdo and Greg Brockman and Sebastien Bubeck and Che Chang and Kai Chen and Mark Chen and Enoch Cheung and Aidan Clark and Dan Cook and Marat Dukhan and Casey Dvorak and Kevin Fives and Vlad Fomenko and Timur Garipov and Kristian Georgiev and Mia Glaese and Tarun Gogineni and Adam Goucher and Lukas Gross and Katia Gil Guzman and John Hallman and Jackie Hehir and Johannes Heidecke and Alec Helyar and Haitang Hu and Romain Huet and Jacob Huh and Saachi Jain and Zach Johnson and Chris Koch and Irina Kofman and Dominik Kundel and Jason Kwon and Volodymyr Kyrylov and Elaine Ya Le and Guillaume Leclerc and James Park Lennon and Scott Lessans and Mario Lezcano-Casado and Yuanzhi Li and Zhuohan Li and Ji Lin and Jordan Liss and Lily and Liu and Jiancheng Liu and Kevin Lu and Chris Lu and Zoran Martinovic and Lindsay McCallum and Josh McGrath and Scott McKinney and Aidan McLaughlin and Song Mei and Steve Mostovoy and Tong Mu and Gideon Myles and Alexander Neitz and Alex Nichol and Jakub Pachocki and Alex Paino and Dana Palmie and Ashley Pantuliano and Giambattista Parascandolo and Jongsoo Park and Leher Pathak and Carolina Paz and Ludovic Peran and Dmitry Pimenov and Michelle Pokrass and Elizabeth Proehl and Huida Qiu and Gaby Raila and Filippo Raso and Hongyu Ren and Kimmy Richardson and David Robinson and Bob Rotsted and Hadi Salman and Suvansh Sanjeev and Max Schwarzer and D. Sculley and Harshit Sikchi and Kendal Simon and Karan Singhal and Yang Song and Dane Stuckey and Zhiqing Sun and Philippe Tillet and Sam Toizer and Foivos Tsimpourlas and Nikhil Vyas and Eric Wallace and Xin Wang and Miles Wang and Olivia Watkins and Kevin Weil and Amy Wendling and Kevin Whinnery and Cedric Whitney and Hannah Wong and Lin Yang and Yu Yang and Michihiro Yasunaga and Kristen Ying and Wojciech Zaremba and Wenting Zhan and Cyril Zhang and Brian Zhang and Eddie Zhang and Shengjia Zhao},
      year={2025},
      eprint={2508.10925},
      archivePrefix={arXiv},
      primaryClass={cs.CL},
      url={https://arxiv.org/abs/2508.10925}, 
}

@misc{openai2024gpt4ocard,
      title={GPT-4o System Card}, 
      author={OpenAI and : and Aaron Hurst and Adam Lerer and Adam P. Goucher and Adam Perelman and Aditya Ramesh and Aidan Clark and AJ Ostrow and Akila Welihinda and Alan Hayes and Alec Radford and Aleksander Mądry and Alex Baker-Whitcomb and Alex Beutel and Alex Borzunov and Alex Carney and Alex Chow and Alex Kirillov and Alex Nichol and Alex Paino and Alex Renzin and Alex Tachard Passos and Alexander Kirillov and Alexi Christakis and Alexis Conneau and Ali Kamali and Allan Jabri and Allison Moyer and Allison Tam and Amadou Crookes and Amin Tootoochian and Amin Tootoonchian and Ananya Kumar and Andrea Vallone and Andrej Karpathy and Andrew Braunstein and Andrew Cann and Andrew Codispoti and Andrew Galu and Andrew Kondrich and Andrew Tulloch and Andrey Mishchenko and Angela Baek and Angela Jiang and Antoine Pelisse and Antonia Woodford and Anuj Gosalia and Arka Dhar and Ashley Pantuliano and Avi Nayak and Avital Oliver and Barret Zoph and Behrooz Ghorbani and Ben Leimberger and Ben Rossen and Ben Sokolowsky and Ben Wang and Benjamin Zweig and Beth Hoover and Blake Samic and Bob McGrew and Bobby Spero and Bogo Giertler and Bowen Cheng and Brad Lightcap and Brandon Walkin and Brendan Quinn and Brian Guarraci and Brian Hsu and Bright Kellogg and Brydon Eastman and Camillo Lugaresi and Carroll Wainwright and Cary Bassin and Cary Hudson and Casey Chu and Chad Nelson and Chak Li and Chan Jun Shern and Channing Conger and Charlotte Barette and Chelsea Voss and Chen Ding and Cheng Lu and Chong Zhang and Chris Beaumont and Chris Hallacy and Chris Koch and Christian Gibson and Christina Kim and Christine Choi and Christine McLeavey and Christopher Hesse and Claudia Fischer and Clemens Winter and Coley Czarnecki and Colin Jarvis and Colin Wei and Constantin Koumouzelis and Dane Sherburn and Daniel Kappler and Daniel Levin and Daniel Levy and David Carr and David Farhi and David Mely and David Robinson and David Sasaki and Denny Jin and Dev Valladares and Dimitris Tsipras and Doug Li and Duc Phong Nguyen and Duncan Findlay and Edede Oiwoh and Edmund Wong and Ehsan Asdar and Elizabeth Proehl and Elizabeth Yang and Eric Antonow and Eric Kramer and Eric Peterson and Eric Sigler and Eric Wallace and Eugene Brevdo and Evan Mays and Farzad Khorasani and Felipe Petroski Such and Filippo Raso and Francis Zhang and Fred von Lohmann and Freddie Sulit and Gabriel Goh and Gene Oden and Geoff Salmon and Giulio Starace and Greg Brockman and Hadi Salman and Haiming Bao and Haitang Hu and Hannah Wong and Haoyu Wang and Heather Schmidt and Heather Whitney and Heewoo Jun and Hendrik Kirchner and Henrique Ponde de Oliveira Pinto and Hongyu Ren and Huiwen Chang and Hyung Won Chung and Ian Kivlichan and Ian O'Connell and Ian O'Connell and Ian Osband and Ian Silber and Ian Sohl and Ibrahim Okuyucu and Ikai Lan and Ilya Kostrikov and Ilya Sutskever and Ingmar Kanitscheider and Ishaan Gulrajani and Jacob Coxon and Jacob Menick and Jakub Pachocki and James Aung and James Betker and James Crooks and James Lennon and Jamie Kiros and Jan Leike and Jane Park and Jason Kwon and Jason Phang and Jason Teplitz and Jason Wei and Jason Wolfe and Jay Chen and Jeff Harris and Jenia Varavva and Jessica Gan Lee and Jessica Shieh and Ji Lin and Jiahui Yu and Jiayi Weng and Jie Tang and Jieqi Yu and Joanne Jang and Joaquin Quinonero Candela and Joe Beutler and Joe Landers and Joel Parish and Johannes Heidecke and John Schulman and Jonathan Lachman and Jonathan McKay and Jonathan Uesato and Jonathan Ward and Jong Wook Kim and Joost Huizinga and Jordan Sitkin and Jos Kraaijeveld and Josh Gross and Josh Kaplan and Josh Snyder and Joshua Achiam and Joy Jiao and Joyce Lee and Juntang Zhuang and Justyn Harriman and Kai Fricke and Kai Hayashi and Karan Singhal and Katy Shi and Kavin Karthik and Kayla Wood and Kendra Rimbach and Kenny Hsu and Kenny Nguyen and Keren Gu-Lemberg and Kevin Button and Kevin Liu and Kiel Howe and Krithika Muthukumar and Kyle Luther and Lama Ahmad and Larry Kai and Lauren Itow and Lauren Workman and Leher Pathak and Leo Chen and Li Jing and Lia Guy and Liam Fedus and Liang Zhou and Lien Mamitsuka and Lilian Weng and Lindsay McCallum and Lindsey Held and Long Ouyang and Louis Feuvrier and Lu Zhang and Lukas Kondraciuk and Lukasz Kaiser and Luke Hewitt and Luke Metz and Lyric Doshi and Mada Aflak and Maddie Simens and Madelaine Boyd and Madeleine Thompson and Marat Dukhan and Mark Chen and Mark Gray and Mark Hudnall and Marvin Zhang and Marwan Aljubeh and Mateusz Litwin and Matthew Zeng and Max Johnson and Maya Shetty and Mayank Gupta and Meghan Shah and Mehmet Yatbaz and Meng Jia Yang and Mengchao Zhong and Mia Glaese and Mianna Chen and Michael Janner and Michael Lampe and Michael Petrov and Michael Wu and Michele Wang and Michelle Fradin and Michelle Pokrass and Miguel Castro and Miguel Oom Temudo de Castro and Mikhail Pavlov and Miles Brundage and Miles Wang and Minal Khan and Mira Murati and Mo Bavarian and Molly Lin and Murat Yesildal and Nacho Soto and Natalia Gimelshein and Natalie Cone and Natalie Staudacher and Natalie Summers and Natan LaFontaine and Neil Chowdhury and Nick Ryder and Nick Stathas and Nick Turley and Nik Tezak and Niko Felix and Nithanth Kudige and Nitish Keskar and Noah Deutsch and Noel Bundick and Nora Puckett and Ofir Nachum and Ola Okelola and Oleg Boiko and Oleg Murk and Oliver Jaffe and Olivia Watkins and Olivier Godement and Owen Campbell-Moore and Patrick Chao and Paul McMillan and Pavel Belov and Peng Su and Peter Bak and Peter Bakkum and Peter Deng and Peter Dolan and Peter Hoeschele and Peter Welinder and Phil Tillet and Philip Pronin and Philippe Tillet and Prafulla Dhariwal and Qiming Yuan and Rachel Dias and Rachel Lim and Rahul Arora and Rajan Troll and Randall Lin and Rapha Gontijo Lopes and Raul Puri and Reah Miyara and Reimar Leike and Renaud Gaubert and Reza Zamani and Ricky Wang and Rob Donnelly and Rob Honsby and Rocky Smith and Rohan Sahai and Rohit Ramchandani and Romain Huet and Rory Carmichael and Rowan Zellers and Roy Chen and Ruby Chen and Ruslan Nigmatullin and Ryan Cheu and Saachi Jain and Sam Altman and Sam Schoenholz and Sam Toizer and Samuel Miserendino and Sandhini Agarwal and Sara Culver and Scott Ethersmith and Scott Gray and Sean Grove and Sean Metzger and Shamez Hermani and Shantanu Jain and Shengjia Zhao and Sherwin Wu and Shino Jomoto and Shirong Wu and Shuaiqi and Xia and Sonia Phene and Spencer Papay and Srinivas Narayanan and Steve Coffey and Steve Lee and Stewart Hall and Suchir Balaji and Tal Broda and Tal Stramer and Tao Xu and Tarun Gogineni and Taya Christianson and Ted Sanders and Tejal Patwardhan and Thomas Cunninghman and Thomas Degry and Thomas Dimson and Thomas Raoux and Thomas Shadwell and Tianhao Zheng and Todd Underwood and Todor Markov and Toki Sherbakov and Tom Rubin and Tom Stasi and Tomer Kaftan and Tristan Heywood and Troy Peterson and Tyce Walters and Tyna Eloundou and Valerie Qi and Veit Moeller and Vinnie Monaco and Vishal Kuo and Vlad Fomenko and Wayne Chang and Weiyi Zheng and Wenda Zhou and Wesam Manassra and Will Sheu and Wojciech Zaremba and Yash Patil and Yilei Qian and Yongjik Kim and Youlong Cheng and Yu Zhang and Yuchen He and Yuchen Zhang and Yujia Jin and Yunxing Dai and Yury Malkov},
      year={2024},
      eprint={2410.21276},
      archivePrefix={arXiv},
      primaryClass={cs.CL},
      url={https://arxiv.org/abs/2410.21276}, 
}
\bibliographystyle{colm2026_conference}

\section*{Reproducibility Statement}

All experiments were conducted on a single NVIDIA H100 GPU.

\textbf{Datasets.} 
We evaluate on \textsc{Geometry3K}~\citep{lu2021intergpsinterpretablegeometryproblem}, 
a benchmark of 3{,}002 multiple-choice plane geometry problems drawn from textbooks, 
and \textsc{PGPS9K}~\citep{zhang2024fusereasonverifygeometry}, an expanded set of 
9{,}022 problems sharing 2{,}891 problems with \textsc{Geometry3K} and adding 
further high-school textbook problems across 4{,}000 unique diagrams. 
Both datasets must be downloaded prior to running the pipeline; 
download instructions are provided in the repository README.

\textbf{Code.}
Our implementation draws major inspiration from~\citet{amanatar2025kaggle} 
for the overall coding setup and inference pipeline structure.
The full source code, including the parallel rollout sampler, entropy 
estimation, sandbox execution, and aggregation pipeline, is available at:
\begin{center}
    \url{https://anonymous.4open.science/r/MARS-GPS-DE55}
\end{center}
Please refer to the \texttt{README} for detailed setup and 
reproduction instructions.

\appendix

\section{Algorithms}
\label{sec:appendix_algorithms}

\begin{algorithm}[H]
\caption{Inference-Time Reasoning with Parallel Rollouts}
\label{alg:reasoning}
\begin{algorithmic}[1]
\REQUIRE Disambiguated formal representation $\mathcal{F}^*$,
         model $f_\theta$, number of rollouts $k$,
         time budget $\mathcal{B}$
\ENSURE Final answer $a^*$

\STATE Construct prompt $\mathcal{P}$ from $\mathcal{F}^*$
\STATE Initialize thread pool (16 workers) and kernel pool
       (16 Jupyter sandboxes)
\FOR{$i = 1$ to $k$ \textbf{in parallel}}
    \STATE $r_i \leftarrow f_\theta(\mathcal{P},\ \tau{=}1.0,\
           \text{min-}p{=}0.02)$ \hfill $\triangleright$ stream
           tokens with logprobs
    \STATE $a_i \leftarrow \textsc{ExtractAnswer}(r_i)$
           \hfill $\triangleright$ parse $\backslash$\texttt{boxed\{N\}}
           or fallback pattern
    \STATE $\bar{H}_i \leftarrow \textsc{MeanEntropy}(r_i)$
           \hfill $\triangleright$ Eq.~\ref{eq:mean_entropy}
    \IF{code block detected in $r_i$}
        \STATE Execute in sandbox; inject output into context
    \ENDIF
\ENDFOR
\STATE $a^* \leftarrow \textsc{AggregateAndVerify}
       (\{a_i\}, \{\bar{H}_i\})$
       \hfill $\triangleright$ Algorithm~\ref{alg:aggregation}
\RETURN $a^*$
\end{algorithmic}
\end{algorithm}

\begin{algorithm}[H]
\caption{Confidence-Aware Answer Aggregation}
\label{alg:aggregation}
\begin{algorithmic}[1]
\REQUIRE Answers $\{a_i\}_{i=1}^k$,
         entropies $\{\bar{H}_i\}_{i=1}^k$,
         model $f_\theta$,
         problem context $\mathcal{F}^*$
\ENSURE Final answer $a^*$

\IF{$\exists\, a : \text{votes}(a) \geq (k/2)+1$}
    \RETURN $a$ \hfill $\triangleright$ Step 1: early consensus
\ENDIF
\IF{$\exists\, a : \text{votes}(a) \geq (k/2)$}
    \RETURN $a$ \hfill $\triangleright$ Step 2: hard accept
\ENDIF

\STATE $\mathcal{A}_{\text{cand}} \leftarrow
       \{a : \text{votes}(a) \geq (k/4)\}$
\IF{$\mathcal{A}_{\text{cand}} = \emptyset$}
    \STATE $\mathcal{A}_{\text{cand}} \leftarrow \{1, 2, 3, 4\}$
\ENDIF
\hfill $\triangleright$ Step 3: candidate selection

\STATE Sort $\mathcal{A}_{\text{cand}}$ by $\bar{H}(a)$ ascending
\hfill $\triangleright$ Step 4: entropy ranking

\FOR{each $a \in \mathcal{A}_{\text{cand}}$ (sorted)}
    \STATE $v \leftarrow f_\theta(\text{VerifyPrompt}(a,\
           \mathcal{F}^*),\ \tau{=}0)$
    \hfill $\triangleright$ Step 5: self-verification
    \IF{$v = \texttt{CORRECT}$}
        \RETURN $a$
    \ENDIF
\ENDFOR

\RETURN $\arg\max_{a}\; \lambda \cdot \text{votes}(a) -
        (1{-}\lambda) \cdot \bar{H}(a)$
\hfill $\triangleright$ Step 6: weighted fallback
\end{algorithmic}
\end{algorithm}

\newpage
\section{CoT Scaling Results}
\label{sec:appendix_full_results}

Table~\ref{tab:cot_scaling} reports accuracy on Geometry3K as a function
of the number of parallel CoT samples $k$.

\begin{table}[H]
\centering
\begin{tabular}{lcc}
\toprule
\textbf{Run} & \textbf{Accuracy (\%)} \\
\midrule
CoT 1   & 82.0 \\
CoT 2   & 85.0 \\
CoT 4   & 86.5 \\
CoT 8   & 87.5 \\
CoT 16  & 88.0 \\
\bottomrule
\end{tabular}
\caption{Accuracy \textit{vs.}\ number of CoT samples on a subset of Geometry3K.}
\label{tab:cot_scaling}
\end{table}

\section{System Prompts}
\label{sec:appendix_prompts}

MARS-GPS uses two system prompts depending on the rollout type. The
\textit{full reasoning prompt} (SP) is used for rollouts requiring
step-by-step chain-of-thought, while the \textit{answer-only prompt}
(AOP) is used for fast silent solves.

\paragraph{Full reasoning prompt (SP).}

\begin{quote}
\small
\texttt{You are an expert geometry problem solver. You will receive a
multiple-choice geometry problem from the Geometry3K benchmark, followed
by structured context automatically extracted by the Pi-GPS parsing
pipeline.}

\medskip
\texttt{INPUT FORMAT}

\texttt{PROBLEM STATEMENT} --- The natural-language geometry question.

\texttt{CHOICES A / B / C / D} --- Four candidate numerical answers.
Exactly one is correct.

\texttt{DIAGRAM LOGIC FORMS} --- First-order predicates automatically
parsed from the diagram image. These are parsed automatically and may
contain minor errors --- treat them as strong hints, not guaranteed truths.

\texttt{TEXT LOGIC FORMS} --- First-order predicates parsed from the
problem text. The first predicate is usually the goal.

\medskip
\texttt{SOLVING PROTOCOL}

\texttt{1.} Read the problem and all four choices.

\texttt{2.} Read the diagram logic forms to understand the figure geometry.

\texttt{3.} Read the text logic forms to confirm the goal and constraints.

\texttt{4.} Solve step by step using the above context.

\texttt{5.} Verify your result against the choices.

\medskip
\texttt{OUTPUT FORMAT}

\texttt{Output ONLY: \textbackslash boxed\{N\}}

\texttt{Where N: 1 = A, 2 = B, 3 = C, 4 = D}

\texttt{Do not write anything after \textbackslash boxed\{N\}.}
\end{quote}

\paragraph{Answer-only prompt (AOP).}

\begin{quote}
\small
\texttt{Geometry MCQ solver. Choices are A/B/C/D. Context given: diagram
logic forms (geometric predicates from the figure), text logic forms
(goal + constraints from problem text). Solve silently. Output only:
\textbackslash boxed\{N\} where 1=A, 2=B, 3=C, 4=D.}
\end{quote}

\section{Worked Example}
\label{sec:appendix_example}

We present a worked example showing the full pipeline for a representative
Geometry3K problem, including the structured context $\mathcal{F}^*$
passed to $f_\theta$ and an abbreviated reasoning trace from one rollout.

\paragraph{Problem.}
\textit{Find PN.}
Choices: A)~25\quad B)~30\quad C)~50\quad D)~60

\paragraph{Structured context $\mathcal{F}^*$.}

\noindent\textbf{Diagram logic forms:}
\begin{quote}
\small\ttfamily
PointLiesOnLine(N, Line(M, C))\\
PointLiesOnLine(L, Line(M, A))\\
Perpendicular(Line(M, L), Line(P, L))\\
Perpendicular(Line(M, N), Line(P, N))\\
Equals(LengthOf(Line(L, P)), 3x+6)\\
Equals(LengthOf(Line(P, N)), 4x-2)\\
Equals(MeasureOf(Angle(B, M, L)), 25)\\
Equals(MeasureOf(Angle(B, M, N)), 25)
\end{quote}

\noindent\textbf{Text logic forms:}
\begin{quote}
\small\ttfamily
Find(LengthOf(Line(P,N)))
\end{quote}

\paragraph{Rollout reasoning trace (abbreviated).}

\begin{quote}
\small
From the diagram, $MP$ bisects $\angle BMC$ since
$\angle BML = \angle BMN = 25°$. By the angle bisector and
perpendicularity conditions, $PL = PN$, so:
\[
3x + 6 = 4x - 2 \implies x = 8
\]

\textit{[calls Python sandbox]}
\begin{verbatim}
x = 8
PN = 4*x - 2
print(PN)
\end{verbatim}
\textit{[sandbox returns: 30]}

\medskip
Therefore $PN = 30$, which matches choice B.

$\backslash$\texttt{boxed\{2\}}
\end{quote}

\noindent\textit{Note: All 8 rollouts agreed on answer B~(30), triggering
early consensus at Step~1. This example illustrates a typical
MARS-GPS execution: the model sets up the algebraic equation,
uses the Python sandbox to evaluate it exactly, and the ensemble
exits immediately via unanimous vote.}

\section{Python Sandbox Ablation Details}
\label{sec:appendix_sandbox}
 
Tables~\ref{tab:sandbox_summary} and~\ref{tab:sandbox_calls} provide
detailed results from the sandbox ablation study described in
Section~\ref{sec:ablation}. The ablation was run on a subset of problems from
Geometry3K with $k{=}8$ rollouts, with $\mathcal{E}$ fully disabled,
code blocks written by $f_\theta$ were intercepted and replaced with a
null response.
 
\begin{table}[H]
\centering
\begin{tabular}{lcc}
\toprule
\textbf{Problem subset} & \textbf{\% of problems} & \textbf{Accuracy} \\
\midrule
All problems                     & 100\%  & 85.0\% \\
Did not attempt Python           &  58.0\% & 92.2\% \\
Attempted Python (sandbox blocked) &  42.0\% & 75.0\% \\
\bottomrule
\end{tabular}
\caption{Accuracy breakdown by whether $f_\theta$ attempted to invoke
the Python sandbox $\mathcal{E}$ during the ablation run. Problems where
$f_\theta$ attempted code execution but was blocked are substantially
harder, with accuracy dropping to 75.0\%.}
\label{tab:sandbox_summary}
\end{table}
 
\begin{table}[H]
\centering
\begin{tabular}{lcc}
\toprule
\textbf{Python calls attempted} & \textbf{\% of problems} & \textbf{Accuracy} \\
\midrule
0 calls   & 58.0\% & 92.2\% \\
1--2 calls & 17.0\% & 79.4\% \\
3--5 calls & 10.0\% & 90.0\% \\
6+ calls   & 15.0\% & 60.0\% \\
\bottomrule
\end{tabular}
\caption{Accuracy by number of Python calls attempted per problem when
the sandbox is disabled. Problems requiring many code calls (6+) drop
to 60.0\% accuracy, suggesting these are the most computationally
intensive cases and benefit most from sandbox access. The recovery in
the 3--5 calls bucket (90.0\%) suggests these problems have sufficient
symbolic structure that $f_\theta$ can partially compensate without
execution.}
\vspace{-5mm}
\label{tab:sandbox_calls}
\end{table}
\section{Execution Time Analysis}
\label{sec:appendix_timing}

Table~\ref{tab:timing_summary} reports average execution time broken down
by prediction outcome, and Table~\ref{tab:timing_ranges} shows how
accuracy varies across execution time ranges. These results are computed
over Geometry3K across two evaluation runs.

\begin{table}[H]
\centering
\begin{tabular}{lcc}
\toprule
\textbf{Outcome} & \textbf{Accuracy} & \textbf{Avg time (s)} \\
\midrule
All problems        & 88.8\% & 47.0  \\
Correct predictions & ---    & 34.6  \\
Wrong predictions   & ---    & 146.1 \\
\bottomrule
\end{tabular}
\caption{Average execution time per problem by prediction outcome.
Wrong predictions consume 4.2$\times$ more time than correct ones,
reflecting the cost of exhausting the verification budget on hard cases.}
\vspace{-5mm}
\label{tab:timing_summary}
\end{table}

\begin{table}[H]
\centering
\begin{tabular}{lcc}
\toprule
\textbf{Time range} & \textbf{\% of problems} & \textbf{Accuracy} \\
\midrule
0--10s    & 35.3\% & 98.6\% \\
10--30s   & 31.2\% & 94.1\% \\
30--60s   & 14.8\% & 89.9\% \\
60--120s  &  9.0\% & 72.2\% \\
120--180s &  3.0\% & 61.1\% \\
180--300s &  3.5\% & 47.6\% \\
$>$300s   &  3.2\% & 42.1\% \\
\bottomrule
\end{tabular}
\caption{Accuracy as a function of execution time range on Geometry3K.
Problems resolved within 10 seconds achieve 98.6\% accuracy, while
problems exceeding 300 seconds drop to 42.1\%, confirming that execution time is a reliable proxy for problem difficulty.}
\vspace{-5mm}
\label{tab:timing_ranges}
\end{table}

\begin{table}[H]
\centering
\begin{tabular}{lcccc}
\toprule
\textbf{Category} & \textbf{Accuracy} &
\textbf{Avg time (s)} & \textbf{Correct (s)} & \textbf{Wrong (s)} \\
\midrule
Similar figures & 96.2\% & 12.9 & 11.0 & 60.0  \\
Trigonometry    & 85.7\% & 12.8 &  5.4 & 57.1  \\
Triangle        & 100\%  & 31.2 & 31.2 & ---   \\
Quadrilateral   & 96.9\% & 34.9 & 28.7 & 227.7 \\
Length/Other    & 85.6\% & 48.3 & 27.1 & 173.6 \\
Angle           & 91.8\% & 51.7 & 45.5 & 121.1 \\
Area            & 77.4\% & 53.3 & 37.8 & 106.5 \\
Circle          & 90.0\% & 56.6 & 42.7 & 181.8 \\
\bottomrule
\end{tabular}
\caption{Per-category accuracy and average execution time on Geometry3K. Area problems have the lowest accuracy (77.4\%) and Circle problems have the highest wrong-prediction time (181.8s), indicating the system spends its full verification budget on these categories before falling back.}
\vspace{-5mm}
\label{tab:timing_category}
\end{table}

\end{document}